# Gibbs Sampling in Open-Universe Stochastic Languages

**Nimar S. Arora**
Computer Science Dept.
University of California
Berkeley, CA 94720

**Rodrigo de Salvo Braz**
Artificial Intelligence Center
SRI International
Menlo Park, CA 94025

**Erik B. Sudderth**
Computer Science Dept.
Brown University
Providence, RI 02912

**Stuart Russell**
Computer Science Dept.
University of California
Berkeley, CA 94720

## Abstract

Languages for open-universe probabilistic models (OUPMs) can represent situations with an unknown number of objects and identity uncertainty. While such cases arise in a wide range of important real-world applications, existing general purpose inference methods for OUPMs are far less efficient than those available for more restricted languages and model classes. This paper goes some way to remedying this deficit by introducing, and proving correct, a generalization of Gibbs sampling to partial worlds with possibly varying model structure. Our approach draws on and extends previous generic OUPM inference methods, as well as auxiliary variable samplers for nonparametric mixture models. It has been implemented for BLOG, a well-known OUPM language. Combined with compile-time optimizations, the resulting algorithm yields very substantial speedups over existing methods on several test cases, and substantially improves the practicality of OUPM languages generally.

## 1 Introduction

General purpose probabilistic modelling languages aim to facilitate the development of complex models while providing effective, general inference methods so that the model-builder need not write model-specific inference code for each application from scratch. For example, BUGS (Spiegelhalter et al., 1996) can represent directed graphical models over indexed sets of random variables and uses MCMC inference (in particular, Gibbs sampling where this is possible).

As the expressive power of modelling languages increases, the range of representable problems also grows. The class of first-order, open-universe probabilistic languages, including BLOG (Milch et al., 2005a) and Church (Goodman et al., 2008), handles cases in which the number of objects (in BUGS, the index set) is unknown and perhaps unbounded, and object identity is uncertain. It is still possible to write a complete inference algorithm for BLOG, based on MCMC over *partial* worlds; each such world is constructed from the minimal self-supporting set of variables relevant to the evidence and query variables. Generality has a price, however: BLOG's default Metropolis–Hastings inference engine samples each variable conditioned only on its parents (Milch & Russell, 2006). This approach leads to unacceptably slow mixing rates for many standard models, in which evidence from child variables is highly informative.

Our goal is to remedy this situation, primarily by extending the range of situations in which Gibbs sampling from the full, conditional posterior can be used within BLOG. Section 2 of this paper introduces the terminology of contingent Bayesian networks (CBNs), which we will use as the propositional "abstract machine" for open-universe stochastic languages. Section 3 surveys previous work related to general purpose sampling of CBNs and describes its limitations. Section 4 then describes our novel Gibbs sampling algorithm for CBNs which addresses these limitations; its implementation for BLOG is described in Section 5. Finally, we present experimental results on various models in Section 6, demonstrating substantial speedups over existing methods.

## 2 Contingent Bayesian Networks

This section repeats, and in some cases generalizes, definitions proposed by Milch et al. (2005b). A *contingent Bayesian network* (CBN) consists of a set of random variables $\mathcal{V}$, and for each variable $X \in \mathcal{V}$, a domain $\text{dom}(X)$ and decision tree $\mathcal{T}_X$. The decision tree is a directed binary tree, where each node is a predi-cate on some subset of $\mathcal{V}$. Each leaf of $\mathcal{T}_X$ encodes a probability distribution parameterized by a subset of $\mathcal{V}$, and defined on $\text{dom}(X)$.

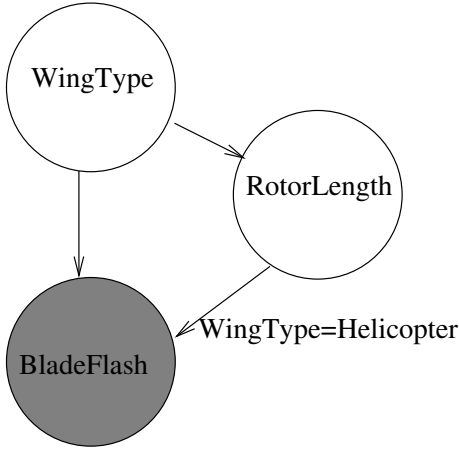

Figure 1: The CBN of Example 1, in which the BladeFlash sensor model differs for helicopters and planes.

**Example 1.** *An aircraft of unknown* WingType – Helicopter *or* FixedWingPlane *– is detected on a radar. Helicopters have an unknown* RotorLength, *and depending on this length they might produce a characteristic pattern called a* BladeFlash *(Tait, 2009) in the returned radar signal. A* FixedWingPlane *might also produce a* BladeFlash. *As summarized in Figure 1,*

$$\mathcal{T}_{\text{WingType}} = F_1$$
$$\mathcal{T}_{\text{RotorLength}} = \begin{cases} F_2 & \text{if WingType} = \text{Helicopter} \\ \text{null} & \text{otherwise} \end{cases}$$
$$\mathcal{T}_{\text{BladeFlash}} = \begin{cases} F_3(\text{RL}) & \text{if WingType} = \text{Helicopter} \\ F_4 & \text{otherwise} \end{cases}$$

*where* RL *is an abbreviation for* RotorLength.

An instantiation $\sigma$ is an assignment of values to a subset of $\mathcal{V}$. We write $\text{vars}(\sigma)$ for the set of variables to which $\sigma$ assigns values, and $\sigma_X$ for the value that $\sigma$ assigns to a variable $X$. $\sigma^{X=a}$ is a modified instantiation which agrees with $\sigma$ except for setting $X$ to $a$. An instantiation $\sigma$ is said to be finite if $\text{vars}(\sigma)$ is finite. An instantiation $\sigma$ *supports* $X$ if all the variables needed to evaluate $\mathcal{T}_X$ are present in $\sigma$. In Example 1, [WingType=FixedWing] supports BladeFlash, but [WingType=Helicopter] does not.

We write $\sigma_{\mathcal{T}_X}$ for the minimal subset of $\sigma$ needed to evaluate $\mathcal{T}_X$, and $p_X(\cdot \mid \sigma_{\mathcal{T}_X})$ for the resulting distribution of $X$. The parents of $X$ in $\sigma$ are $\text{vars}(\sigma_{\mathcal{T}_X})$, while the children of $X$ in $\sigma$ are

$$\Lambda(\sigma, X) = \{Y \mid Y \in \text{vars}(\sigma), X \in \text{vars}(\sigma_{\mathcal{T}_Y})\}. \quad (1)$$

The subset of $\text{vars}(\sigma_{\mathcal{T}_X})$ which were used to evaluate internal nodes of $\mathcal{T}_X$ (rather than the leaf) are the *switching parents* of $X$ in $\sigma$. Intuitively, changes in the values of switching parents can switch the distribution of $X$, as well as its set of parents. A *switching variable* in $\sigma$ is a switching parent for one or more variables in $\sigma$. For the CBN of Example 1, the instantiation [ WingType=Helicopter, RotorLength=6, BladeFlash=true ] makes WingType a switching parent of both RotorLength and BladeFlash.

An instantiation $\sigma$ is *self-supporting* if it supports all variables in $\sigma$. Assuming that the CBN is well-defined (Milch et al., 2005b), we can define the probability of a self-supporting instantiation as follows:

$$p(\sigma) = \prod_{X \in \text{vars}(\sigma)} p_X(\sigma_X \mid \sigma_{\mathcal{T}_X}) \quad (2)$$

An instantiation $\sigma$ is *feasible* if $p(\sigma) > 0$.

## 3 Related Work

Milch and Russell (2006) have previously shown that the state space for *Markov chain Monte Carlo* (MCMC) inference in CBNs may consist of minimal partial instantiations that support the evidence, $E$, and query variables, $Q$. This idea has been exploited to build the current, default inference engine for BLOG. Standard sampling algorithms for nonparametric, Dirichlet process mixture models use a related representation: they instantiate parameters for those mixture components which support the evidence, as well as a few auxiliary components (Neal, 2000). Our new algorithm builds on both of these methods.

### 3.1 Parent-Conditional Sampling

In the absence of a model-specific, user supplied proposal distribution, BLOG's existing inference engine relies on a *parent-conditional* proposal. This algorithm picks a variable, $X$, at random from all non-evidence variables in the current instantiation $\sigma$, $V(\sigma) = \text{vars}(\sigma) - E$, and proposes a new instantiation $\sigma'$ with the value of $X$ drawn from $p_X(\cdot \mid \sigma_{\mathcal{T}_X})$. If $X$ was a switching variable in $\sigma$, we may then need to instantiate new variables, and uninstantiate unneeded ones, to make $\sigma'$ minimal and self-supporting over $Q \cup E$. All new variables are instantiated with values drawn from their parent-conditional distribution.

We say that any $\sigma'$ constructed by this procedure is *reachable* from $\sigma$ via $X$, or $\sigma \overset{X}{\leadsto} \sigma'$. The following properties are easily seen to be true of reachability.

**Proposition 1.** *A minimal self-supporting feasible instantiation $\sigma'$ is reachable from an instantiation $\sigma$ via $X$ if and only if $X \in \text{vars}(\sigma) \cap \text{vars}(\sigma')$, and $\sigma$ and $\sigma'$ agree on all other variables in $\text{vars}(\sigma) \cap \text{vars}(\sigma')$.*

**Proposition 2.** *If $\sigma \overset{X}{\leadsto} \sigma'$, then there does not exist $Y \in V(\sigma)$, $Y \neq X$, such that $\sigma \overset{Y}{\leadsto} \sigma'$.*

The nature of this proposal distribution $q(\sigma \to \sigma')$ makes it quite simple to compute the acceptance ratio for the Metropolis–Hastings (MH) method (Andrieu et al., 2003), which takes the following form:

$$\alpha(\sigma \to \sigma') = \min\left\{1, \frac{p(\sigma')q(\sigma' \to \sigma)}{p(\sigma)q(\sigma \to \sigma')}\right\} \quad (3)$$

For any $\sigma'$ reachable from $\sigma$ via $X$, the unique way of proposing this transition is to select $X$ from $V(\sigma)$, propose the value $\sigma'_X$ for it, and finally propose corresponding values for all new variables in $\sigma'$. Thus,

$$q(\sigma \to \sigma') = \frac{p_X(\sigma'_X \mid \sigma_{\mathcal{T}_X})}{|V(\sigma)|} \prod_{Y \in \text{vars}(\sigma') - \text{vars}(\sigma)} p_Y(\sigma'_Y \mid \sigma'_{\mathcal{T}_Y}) \quad (4)$$

From Equations (2) and (4), the terms corresponding to $\text{vars}(\sigma') - \text{vars}(\sigma)$ cancel in $p(\sigma')/q(\sigma \to \sigma')$. Similarly, terms in $\text{vars}(\sigma) - \text{vars}(\sigma')$ cancel in $q(\sigma' \to \sigma)/p(\sigma)$. Further, it is easy to see that for variables $Y \in \text{vars}(\sigma) \cap \text{vars}(\sigma') - \Lambda(\sigma, X) \cap \Lambda(\sigma', X)$, $\sigma_{\mathcal{T}_Y} = \sigma'_{\mathcal{T}_Y}$. Hence, $p_Y(\cdot \mid \sigma_{\mathcal{T}_Y}) = p_Y(\cdot \mid \sigma'_{\mathcal{T}_Y})$ and the terms for all such variables $Y$, including $X$, cancel out. Finally, the acceptance ratio $\alpha(\sigma \to \sigma')$ reduces to:

$$\min\left\{1, \frac{|V(\sigma)|}{|V(\sigma')|} \prod_{Y \in \Lambda(\sigma, X) \cap \Lambda(\sigma', X)} \frac{p_Y(\sigma'_Y \mid \sigma'_{\mathcal{T}_Y})}{p_Y(\sigma_Y \mid \sigma_{\mathcal{T}_Y})}\right\} \quad (5)$$

Note the dependence on those variables which are children of $X$ in both $\sigma$ and $\sigma'$. The overall algorithm is summarized in Figure 2.

### 3.2 Gibbs Sampling

Equation (5) summarizes the main problem with parent-conditional sampling: if the proposed value for the sampled variable $X$ does not assign high probability to the children of $X$, the move will be rejected. To avoid undue assumptions, hierarchical Bayesian statistical models often use dispersed or "vague" priors, so that such parent-conditional proposals have extremely low acceptance probabilities.

The *Gibbs sampler* addresses this issue by directly sampling $X$ from its *full* conditional distribution, $p_X(\cdot \mid \sigma_{\mathcal{V}-X})$, rather than its parent-conditional prior $p_X(\cdot \mid \sigma_{\mathcal{T}_X})$. This method was originally proposed by Geman and Geman (1984) for inference in undirected Markov random fields, and later popularized as a general Bayesian inference method by Gelfand and Smith (1990). For discrete variables $X$, the Gibbs sampler computes a weight $w(a)$ for each $a \in \text{dom}(X)$:

$$w(a) = p_X(a \mid \sigma_{\mathcal{T}_X}) \prod_{Y \in \Lambda(\sigma, X)} p_Y(\sigma_Y \mid \sigma_{\mathcal{T}_Y}^{X=a}) \quad (6)$$

1. Create an initial, minimal, self-supporting feasible instantiation $\sigma$ consistent with the evidence $E$, and including the query variables $Q$.

2. Initialize statistics of the query variables to zero.

3. Repeat for the desired number of iterations:
   (a) Choose $X \in V(\sigma)$ uniformly at random.
   (b) Randomly propose $\sigma'$ such that $\sigma \overset{X}{\leadsto} \sigma'$ using the distribution of Equation (4).
   (c) Compute the acceptance ratio, $\alpha(\sigma \to \sigma')$, via Equation (5).
   (d) With probability $\alpha(\sigma \to \sigma')$, set $\sigma \leftarrow \sigma'$. Otherwise, leave $\sigma$ unchanged.
   (e) Update query statistics using $\sigma$.

4. Report query statistics.

Figure 2: General purpose inference in CBNs using parent-conditional Metropolis–Hastings proposals, as in (Milch & Russell, 2006).

A new value $\sigma'_X$ is then sampled from a normalized distribution with mass proportional to these non-negative weights. Viewed as a Metropolis-Hastings proposal, the acceptance probability for the Gibbs sampler always equals one; Gibbs moves are never rejected.

The Gibbs sampler can be consistently applied to variables with finite, countable, or even uncountable domains, so long as the full conditional posterior can be tractably normalized and sampled from. For models specified via languages like BUGS, Gibbs sampling has proven quite successful. However, most existing applications and analysis of the Gibbs sampler implicitly assume a closed universe model, and instantiate the full, finite set of variables at all iterations. If this algorithm were naively applied to a CBN, then for some switching variables $X$ and configurations $a \in \text{dom}(X)$, $\sigma^{X=a}$ might not support some children of $X$. For such inconsistent model configurations, the normal Gibbs weight $w(a)$ cannot be evaluated.

One possible solution, proposed in the context of Dirichlet process (DP) mixture models by Neal (2000), augments $\sigma$ with *auxiliary variables* chosen so that $\sigma^{X=a}$ is self-supporting for all $a \in \text{dom}(X)$. This augmented $\sigma$, which is now no longer minimal, is used to construct the Gibbs weights; following the move any remaining non-supported variables are discarded.

dom($X$) = {0, 1, 2}
$X \sim$ Categorical(.1, .6, .3)

dom($Y_i$) = {0, 1} for all $i \in \mathcal{N}$
$Y_i \sim \begin{cases} \text{Bernoulli}(\frac{1}{1+X}) & \text{if } (X+i) \mod 2 \equiv 0 \\ \text{Bernoulli}(\frac{1}{1+X+Y_{i+1}}) & \text{otherwise} \end{cases}$

Evidence: $Y_1$ = true. Query: $X$.

Figure 3: A CBN which requires infinitely many auxiliary variables for standard Gibbs sampling approaches.

Such auxiliary variables are always sampled conditioned on $\sigma$, given the current value of $X$. For example, if $\sigma_X = a$ and if $\sigma$ was augmented with a variable $Z$ needed to support $\sigma^{X=b}$ for some $b \in \text{dom}(X) - a$, then we would sample $Z$ from $p_Z(\cdot \mid \sigma_{\mathcal{T}_Z}^{X=a})$. This can lead to poor mixing rates, or an inconsistent sampler if $p_Z(\cdot \mid \sigma_{\mathcal{T}_Z}^{X=a})$ and $p_Z(\cdot \mid \sigma_{\mathcal{T}_Z}^{X=b})$ have non-overlapping support. Note that this issue doesn't arise with the DP mixture sampler, since $\mathcal{T}_Z$ had no dependence on $X$, and $p_Z(\cdot \mid \sigma_{\mathcal{T}_Z}^{X=a}) = p_Z(\cdot \mid \sigma_{\mathcal{T}_Z}^{X=b})$ for any $a, b$.

To further illustrate this issue, consider the model of Example 1 and a minimal instantiation, $\sigma$ = [ WingType = FixedWingPlane, BladeFlash = True ]. If we were to apply a typical auxiliary variable method to do MCMC sampling in this model, we would first instantiate RotorLength given WingType = FixedWingPlane, and then construct Gibbs weights for WingType = FixedWingPlane and Helicopter. However, the only value of RotorLength that can be sampled given WingType = FixedWingPlane is null, and this value has probability 0 with WingType = Helicopter. The resulting chain will *not* mix to the true posterior.

In fact, there are cases when the auxiliary variable method is not well defined, because we may need an unbounded number of auxiliary variables. Consider the rather artificial but instructive CBN in Figure 3, and an instantiation $\sigma = [X = 0, Y_1 = 1, Y_2 = 1]$. To augment $\sigma$ such that it is self-supporting for all values of $X$, we certainly need to add $Y_3$, since $Y_2$ depends on $Y_3$ when $X = 1$. But $Y_3$ depends on $Y_4$ when $X = 0$, and so we need to add $Y_4$, and so on. Ultimately, we would need to instantiate $Y_i$ for all $i \geq 1$.

## 4 Gibbs Sampling in Contingent Bayesian Networks

We now develop a general-purpose extension of standard Gibbs samplers, which is applicable to arbitrary switching variables with finite domains. The proposal for a switching variable, $X$, will proceed in three steps. First, the instantiation, $\sigma$, is reduced to a subset of variables, core($\sigma, X$), that is guaran-teed to exist in a minimal, self-supporting instantiation constructed from $\sigma^{X=a}$, for any $a \in dom(X)$. Second, we construct minimal self-supporting instantiations $\sigma_i$, $i = 1, \ldots, |\text{dom}(X)| - 1$, for each value in $\text{dom}(X) - \{\sigma_X\}$. These instantiations agree with $\sigma$ on core($\sigma, X$), but assign different values to $X$. Any remaining variables in these $\sigma_i$ configurations are sampled from their parent-conditional priors. For notational simplicity, we define $\sigma_0 = \sigma$. Finally, we assign weights to these $\sigma_i$, $i = 0, \ldots, |\text{dom}(X)| - 1$, and make a transition proportional to these weights.

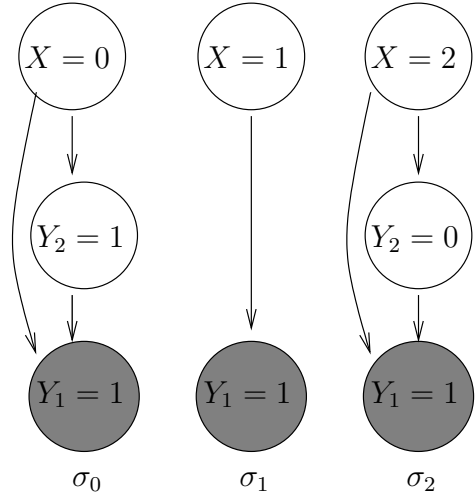

Figure 4: The three partial instantiations considered for Gibbs sampling of $X$ given $Y_1$ as evidence. Here, $\sigma_0$ is the current instantiation, and $\sigma_1$, $\sigma_2$ are candidate new configurations.

It may seem counter-intuitive to first reduce the instantiation, and then extend it. After all, the pair of algorithms described in Section 3, parent-conditional sampling and auxiliary variable Gibbs sampling, first extended the current instantiation before reducing it. The motivation for our approach is simple: variables whose existence depends on the value of $X$ should be sampled in a world with the appropriate value of $X$.

Consider again, for example, the model in Figure 3, and three partial instantiations illustrated in Figure 4. Now, starting from $\sigma_0$ (in which $X = 0$), we could have fixed the value of $Y_2$ when constructing $\sigma_2$ (in which $X = 2$). However, the distribution of $Y_2$ given $X = 2$ is quite different from that given $X = 0$, and fixing the value of $Y_2$ could lead to low probability instantiations. The resampling of non-core variables like $Y_2$ also simplifies the detailed balance equations discussed later. In particular, our algorithm is designed so that the distribution of $\sigma_2$ does not depend on whether we start from $\sigma_0$ or $\sigma_1$. Thus, when demonstrating detailed balance between pairs of instantiations, we need not reason about other instantiations which might be involved in the transition. This last observation relies on the fact that $\text{core}(\sigma_0, X) = \text{core}(\sigma_1, X)$. We will first prove this in general.

**Definition 1.** *For an instantiation $\sigma$ and variables $X, Y, Z \in \text{vars}(\sigma)$, if $\mathcal{T}_Z$ refers to $X$ and $Y$, and the first reference to $X$ precedes the first reference to $Y$, the edge linking $Y$ to $Z$ is said to be* contingent *on $X$.*

**Definition 2.** *Let $\text{core}(\sigma, X)$ denote the subset of variables in $\text{vars}(\sigma) - \{X\}$ which have a path (possibly of length zero) consisting of parent-child edges, excluding edges contingent on $X$, to some variable in $Q \cup E$.*

Note that we have left $X$ out of $\text{core}(\sigma, X)$ mainly for simplifying the subsequent text. However, it is not hard to see that there is a path from $X$ to $Q \cup E$ not contingent upon $X$. For example, consider the shortest path from $X$ to $Q \cup E$ and let this path start with the $X \to Y$ edge. Now, the edge $X \to Y$ is not contingent upon $X$ (by definition) and if some other edge, $W \to Z$, along this path is contingent upon $X$ then we can find a shorter path starting with $X \to Z$. It should be further noted that all the ancestors of $X$ have a path to $X$ not contingent upon $X$ (otherwise, a cyclic instantiation would make the CBN not well-formed). Hence all the ancestors of $X$ are in $\text{core}(\sigma, X)$.

**Definition 3.** *For an instantiation $\sigma$ and variable $X \in \text{vars}(\sigma)$, let $\Upsilon(\sigma, X) \triangleq \Lambda(\sigma, X) \cap \text{core}(\sigma, X)$ denote the children of $X$ also contained in $\text{core}(\sigma, X)$.*

**Proposition 3.** *For any pair of minimal self-supporting instantiations, $\sigma$ and $\sigma'$, and variable $X$ common to them, if $\sigma$ and $\sigma'$ agree on $\text{core}(\sigma, X)$ then $\text{core}(\sigma, X) = \text{core}(\sigma', X)$ and $\Upsilon(\sigma, X) = \Upsilon(\sigma', X)$.*

*Proof.* Let $Y \in \text{core}(\sigma, X)$, then either $Y \in Q \cup E$ or there exists a path of edges not contingent on $X$ from $Y$ to $Q \cup E$. Clearly, if $Y \in Q \cup E$ then $Y \in \text{core}(\sigma', X)$. Otherwise, let $Z$ be the first child in such a path. Since $X$ is not referenced before $Y$ in $\mathcal{T}_Z$, $X$ is also not referenced before any $W$ referenced before $Y$ in $\mathcal{T}_Z$. Such a variable $W$ must also be in $\text{core}(\sigma, X)$ since $W$ has the same path to $Q \cup E$ via $Z$ as $Y$. But $\sigma$ and $\sigma'$ agree on $\text{core}(\sigma, X)$ and hence on $W$. Since $\sigma$ and $\sigma'$ agree on all the variables referred before $Y$ in $\mathcal{T}_Z$ it follows that the evaluation of $\mathcal{T}_Z$ up to $Y$ is identical in $\sigma$ and $\sigma'$. Hence, the $Y$ to $Z$ edge is not contingent on $X$ in $\sigma'$. By induction, the path from $Y$ to $Q \cup E$ in $\sigma'$ is not contingent on $X$, which implies that $Y \in \text{core}(\sigma', X)$.

Now, suppose $\text{core}(\sigma, X) \subset \text{core}(\sigma', X)$. For any element in $\text{core}(\sigma', X) - \text{core}(\sigma, X)$ there must be a path of edges not contingent upon $X$ in $\sigma'$ to $Q \cup E$ via some variables in $\text{core}(\sigma, X) \cup \{X\}$ (trivially, since $Q \cup E \subseteq \text{core}(\sigma, X) \cup \{X\}$). Let $Y$ and $Z$ be one such parent-child pair in $\sigma'$ s.t. $Y \in \text{core}(\sigma', X) - \text{core}(\sigma, X)$ and $Z \in \text{core}(\sigma, X) \cup \{X\}$. Now, all the variables referred in $\mathcal{T}_Z$ up to the first reference of $X$ (if any) would also be in $\text{core}(\sigma, X)$ since they have an edge to $Z$ which is not contingent on $X$. Since $\sigma$ and $\sigma'$ agree on $\text{core}(\sigma, X)$, the evaluation of $\mathcal{T}_Z$ would follow an identical path in $\sigma$ and $\sigma'$ up to the first reference of $X$. Therefore, since $Y$ is not referred to after $X$ while evaluating $\mathcal{T}_Z$ in $\sigma'$, it follows that $Y \in \text{core}(\sigma, X)$.

Let $Y \in \Upsilon(\sigma, X)$, i.e. $Y$ is a child of $X$ in $\sigma$ and $Y \in \text{core}(\sigma, X)$. From the above result $Y \in \text{core}(\sigma', X)$ and we will next show that $Y$ is a child of $X$ in $\sigma'$. Consider the evaluation path of $\mathcal{T}_Y$ in $\sigma$. All the variables that are referred before $X$ are also in $\text{core}(\sigma, X)$ by definition. Since these variables will have the same value in $\sigma'$, it follows that the evaluation of $\mathcal{T}_Y$ in $\sigma'$ will lead to $X$ being referred. In other words, $X$ is a parent of $Y$ in $\sigma'$ which implies that $\Upsilon(\sigma, X) \subseteq \Upsilon(\sigma', X)$. By symmetry, $\Upsilon(\sigma', X) \subseteq \Upsilon(\sigma, X)$ □

**Proposition 4.** *For any two minimal self-supporting instantiations, $\sigma$ and $\sigma'$, there is at most one variable $X$ common to them such that $\sigma$ and $\sigma'$ agree on $\text{core}(\sigma, X)$, but differ on $X$.*

*Proof.* Assume to the contrary that there exist two such variables $X$ and $Y$. Now, since $\sigma$ and $\sigma'$ agree on $\text{core}(\sigma, X)$ but differ on $Y$, it follows that $Y \notin \text{core}(\sigma, X)$. Hence $Y$ cannot be in $Q \cup E$. But since $\sigma$ is a minimal instantiation, $Y$ must have a path to $Q \cup E$. Now consider the shortest path of $Y$ to $Q \cup E$. Some edge, $W \to Z$, in this path must be contingent on $X$. Hence we can construct a path from $X$ to $Q \cup E$ via $Z$ which can't be contingent on $Y$ (otherwise, $Y$ would have a shorter path to $Q \cup E$). This implies that $X \in \text{core}(\sigma, Y)$, but then $\sigma$ and $\sigma'$ agree on $X$, a contradiction. □

For each value in $\text{dom}(X)$, the corresponding partial instantiation $\sigma_i$ is assigned the following weight:

$$w(\sigma_i) = \frac{p_X(\sigma_{iX} \mid \sigma_{i\mathcal{T}_X})}{|V(\sigma_i)|} \prod_{Y \in \Upsilon(\sigma, X)} p_Y(\sigma_{iY} \mid \sigma_{i\mathcal{T}_Y}) \tag{7}$$

Up to a multiplicative constant, this expression reduces to Equation 6 if $X$ is not a switching variable. The complete pseudo-code is given in Figure 5. Note that if $X$ is not a switching variable then $core(\sigma, X) = vars(\sigma) - X$ and the algorithm reduces to regular Gibbs sampling.

It only remains to show that detailed balance holds between any two minimal instantiations $\sigma_0$ and $\sigma_1$. It follows from Propositions 3 and 4 that there is at most one shared variable $X$ such that a transition is possible between $\sigma_0$ and $\sigma_1$ by sampling $X$. Thus, the only way for this transition to occur from

1. Create an initial, minimal, self-supporting feasible instantiation $\sigma$ consistent with the evidence $E$, and including the query variables $Q$.

2. Initialize statistics of the query variables to zero.

3. Repeat for the desired number of iterations:

   (a) Choose $X \in V(\sigma)$ uniformly at random.

   (b) If $X$ has finite domain (say, $dom(X) = \{v_0, \ldots, v_{n-1}\}$ and $\sigma_X = v_0$).

      i. Compute $core(\sigma, X)$.

      ii. Construct $\sigma_i$: $core(\sigma, X) \cup \{X\} \overset{X=v_i}{\leadsto} \sigma_i$ for $i = 1, \ldots, n-1$.

      iii. Compute $w(\sigma_i)$ from Equation (7) for $i = 0, \ldots, n-1$.

      iv. Normalize $w(\cdot)$ and sample an index $j$ from this distribution. Set $\sigma \leftarrow \sigma_j$.

      Else

      i. Randomly propose $\sigma'$ such that $\sigma \overset{X}{\leadsto} \sigma'$ using the distribution of Equation (4).

      ii. Compute the acceptance ratio, $\alpha(\sigma \to \sigma')$, via Equation (5).

      iii. With probability $\alpha(\sigma \to \sigma')$, set $\sigma \leftarrow \sigma'$. Otherwise, leave $\sigma$ unchanged.

   (c) Update query statistics using $\sigma$.

4. Report query statistics.

Figure 5: General purpose Gibbs sampling in CBNs

$\sigma_0$ is to first select $X$ for sampling with probability $\frac{1}{|V(\sigma_0)|}$. Next, the new variables in $\sigma_1$, $\psi(\sigma_0, X, \sigma_1) = vars(\sigma_1) - core(\sigma_0, X) - \{X\}$, must be sampled with probability $\prod_{Y \in \psi(\sigma_0, X, \sigma_1)} p_Y(\sigma_{1Y}|\sigma_{1\mathcal{T}_Y})$. Finally, we must select $\sigma_1$ out of all the other random instantiations, with probability $\frac{w(\sigma_1)}{w(\sigma_0)+\ldots+w(\sigma_{n-1})}$. Now, the instantiations $\sigma_2, \ldots, \sigma_{n-1}$ are random variables and hence the overall transition probability, $q(\sigma_0 \to \sigma_1)$, depends on the expected value of this last probability under the distribution of these random variables:

$$\frac{1}{|V(\sigma_0)|} \prod_{Y \in \psi(\sigma_0, X, \sigma_1)} p_Y(\sigma_{1Y} \mid \sigma_{1\mathcal{T}_Y}) E\left[\frac{w(\sigma_1)}{\sum_{i=0}^{n-1} w(\sigma_i)}\right]$$

We can construct a similar expression for the reverse move probability, and note that the numerator in the expectation is a constant, and the rest of the expectation doesn't depend on which of $\sigma_0$ or $\sigma_1$ we start out with. Thus, $\frac{q(\sigma_0 \to \sigma_1)}{q(\sigma_1 \to \sigma_0)}$ is:

$$\frac{|V(\sigma_1)|}{|V(\sigma_0)|} \frac{w(\sigma_1)}{w(\sigma_0)} \frac{\prod_{Y \in \psi(\sigma_0, X, \sigma_1)} p_Y(\sigma_{1Y} \mid \sigma_{1\mathcal{T}_Y})}{\prod_{Y \in \psi(\sigma_1, X, \sigma_0)} p_Y(\sigma_{0Y} \mid \sigma_{0\mathcal{T}_Y})}$$

Substituting for $w(\sigma_1)$ and $w(\sigma_0)$:

$$\frac{q(\sigma_0 \to \sigma_1)}{q(\sigma_1 \to \sigma_0)} = \prod_{Y \in \Upsilon(\sigma, X)} \frac{p_Y(\sigma_{1Y} \mid \sigma_{1\mathcal{T}_Y})}{p_Y(\sigma_{0Y} \mid \sigma_{0\mathcal{T}_Y})}$$

$$\frac{p_X(\sigma_{1X} \mid \sigma_{1\mathcal{T}_X})}{p_X(\sigma_{0X} \mid \sigma_{0\mathcal{T}_X})} \cdot \frac{\prod_{Y \in \psi(\sigma_0, X, \sigma_1)} p_Y(\sigma_{1Y} \mid \sigma_{1\mathcal{T}_Y})}{\prod_{Y \in \psi(\sigma_1, X, \sigma_0)} p_Y(\sigma_{0Y} \mid \sigma_{0\mathcal{T}_Y})}$$

Observe that the only terms missing from $\frac{p(\sigma_1)}{p(\sigma_0)}$ above are those for variables in $core(\sigma, X) - \Upsilon(\sigma, X)$. However, if $Y \in core(\sigma, X)$ then $\sigma_Y = \sigma'_Y$ and further if $Y \notin \Lambda(\sigma, X)$ all the parents of $Y$ are also in $core(\sigma, X)$ and hence have the same values in $\sigma$ and $\sigma'$. Thus these variables have identical values and distributions in $\sigma_0$ and $\sigma_1$ and their terms cancel out. Finally,

$$\frac{q(\sigma_0 \to \sigma_1)}{q(\sigma_1 \to \sigma_0)} = \frac{p(\sigma_1)}{p(\sigma_0)}$$

## 5 BLOG Compiler

We have implemented our algorithm in a new implementation of the BLOG language, which we will refer to as *blogc*[1]. The broad outline of our implementation is similar to Milch's public-domain Metropolis-Hastings version, except in two significant aspects.

First, for variables with (possibly unknown) finite domain, we always use Gibbs sampling. By statically analyzing the structure of the model we can determine which variables are switching variables, which ones need to be resampled for each transition, etc. Based on the analysis, appropriate code is generated that does the actual sampling and reporting.

Consider, as an example, the BLOG model in Figure 6. This model describes the prior distribution of two types of aircraft – fixed-wing planes and helicopters. These planes may produce an arbitrary number of blips on the radar (the fact that plane $a$ produces a blip $b$ is represented by setting Source$(b) = a$). Further, helicopters due to the interaction of their rotor with the radar beam can produce blade-flashes in the radar blip. In this model, the variable RotorLength$(a)$ for all aircraft $a$ can easily be Gibbs sampled. If WingType$(a)$ =Helicopter then RotorLength$(a)$ can be either Short or Long, otherwise it can only be null (as per BLOG semantics for a missing else clause). While compiling the model we can detect that the children variables of WingType$(a)$ in any instantiation are all the BladeFlash$(b)$ variables such that Source$(b) = a$. In order to speed up the computation of the Gibbs weights at runtime, we maintain a list, for each object $a$ of type Aircraft, of all objects $b$ of type Blip such that Source$(b) = a$.

---

[1] *blogc* is available for download from: http://code.google.com/p/blogc/

The variable WingType($a$) is more interesting. It can only take two possible values, but since it is a switching variable, care has to be taken when sampling it. In particular, the variable RotorLength($a$) has to be uninstantiated. This is because all the children edges from RotorLength($a$) are contingent on the value of WingType($a$). Note that Source($b$) for all objects $b$ of type Blip is also a switching variable. However, in this case the decision to uninstantiate a variable WingType($a$) such that Source($b$) = $a$ depends on whether there exists another object $b'$ such that Source($b'$) = $a$.

The second major difference in our implementation is the handling of number variables. Instead of directly sampling the number variables, our implementation proposes birth and death moves. In the radar example, for each object $w$ of type WingType, we generate an Aircraft object that has no blips assigned to it. The death move kills off such objects with no blips. In order to get faster mixing, we allow some extra flexibility in the birth and death move during an "initialization" phase. During this phase, birth and death moves ignore the probability of child variables. To understand the motivation, assume for a moment that the expected number of blips for a given aircraft was one million. Now, a birth move which proposes an aircraft with 0 blips would be almost certainly rejected. By allowing such birth moves during initialization, we give the inference engine an opportunity to later attach blips to the aircraft.

## 6 Experimental Results

We have compared the convergence speed and accuracy of blogc against the existing generic Metropolis-Hastings inference engine provided with BLOG, which we will refer to as BLOG-MH. Since a Gibbs and a MH sampler perform different amount of work in each sample we felt that it was more appropriate to compare the two inference engines with respect to time. In order to control for the compiler optimizations in blogc we have implemented a version of BLOG-MH in blogc which we will refer to as *blogc-MH*. For some of the other experiments we have also implemented a version of Gibbs sampling that doesn't uninstantiate and resample variables not in the core, which we shall refer to as *blogc-noblock*.

In the following three models each inference engine is run for a varying number of samples, where a sample is as defined by that inference engine. For each number of samples, inference is repeated 20 times with a different random seed and the mean and variance of a query variable is plotted against the average elapsed time (in seconds).

```
type AircraftType;
type Length;
type Aircraft;
type Blip;

origin AircraftType WingType(Aircraft);
random Length RotorLength(Aircraft);
origin Aircraft Source(Blip);
random Boolean BladeFlash(Blip);

guaranteed AircraftType Helicopter, FixedWingPlane;
guaranteed Length Short, Long;

#Aircraft(WingType = w)
  if w = Helicopter then
    ~Poisson [1.0]
  else
    ~Poisson [4.0];

#Blip ~Poisson[2.0];

#Blip(Source = a) ~ Poisson[1.0];

RotorLength(a) {
  if WingType(a) = Helicopter then
    ~TabularCPD [[0.4, 0.6]]
};

BladeFlash(b) {
  if Source(b) = null then
    ~Bernoulli [.01]
  elseif WingType(Source(b)) = Helicopter then
    ~TabularCPD[[.9,.1],[.6,.4]]
            (RotorLength(Source(b)))
  else
    ~Bernoulli [.1]
};

obs {Blip b} = {b1, b2, b3, b4, b5, b6};

obs BladeFlash(b1) = true;
obs BladeFlash(b2) = false;
obs BladeFlash(b3) = false;
obs BladeFlash(b4) = false;
obs BladeFlash(b5) = false;
obs BladeFlash(b6) = false;

query WingType(Source(b1));
query WingType(Source(b2));
query WingType(Source(b3));
query WingType(Source(b4));
query WingType(Source(b5));
query WingType(Source(b6));
```

Figure 6: Example of helicopters and fixed-wing planes being detected by a radar

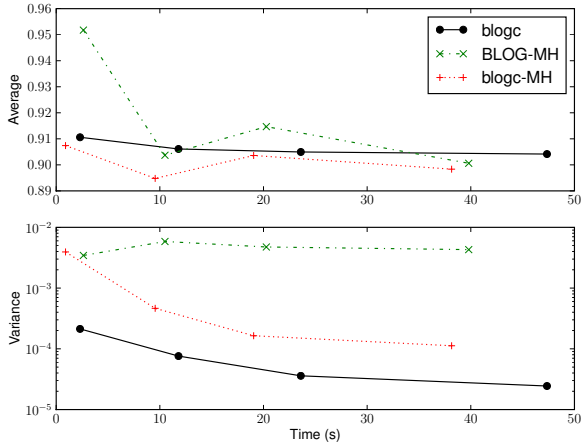

Figure 7: Results on the Alarm Bayes Net

First, we evaluate on the Alarm network of (Beinlich et al., 1989) available from the Bayes Network Repository[2] (Friedman et al., 1997). This is a Bayes Net with 37 discrete random variables of which we observe 9. The results are summarized in Figure 7. The important thing to note is that the variance achieved by blogc in less than 2 seconds is much better than that achieved by blogc-MH in 15 seconds and by BLOG-MH in 40 seconds. The compiler optimizations are clearly giving a big boost but the Gibbs sampling is helping considerably as well.

Next, we consider the model in Figure 8 which is the urns-and-balls example of (Milch et al., 2005a) with a slight twist. Balls have a weight instead of a discrete color. Figure 9 shows that blogc converges significantly faster than BLOG-MH. However, all the improvement here is being driven by the compiler optimizations as evidenced by the fact that blogc-MH is keeping pace with blogc. This similarity is likely due to the fact that our current blogc implementation does not resample the TrueWeight variables from their full posterior. This shortcoming arises because blogc does not yet support Gibbs updates for continuous variables, and is not a limitation of the proposed Gibbs sampler for switching variables. Nevertheless, the example demonstrates the soundness of the blogc-MH implementation in addition to the compiler optimizations.

Our final result is on the radar example of Figure 6. For this model we experimented running blogc without the logic which detects that $RotorLen(a)$ must be uninstantiated when sampling $WingType(a)$. This mode is labeled as blogc-noblock in Figure 10. In this experiment we are querying the probability that

---

[2]http://www.cs.huji.ac.il/site/labs/compbio/Repository/

```
type Ball;
type Draw;

random Real TrueWeight(Ball);
random Ball BallDrawn(Draw);
random Real ObsWeight(Draw);

guaranteed Draw Draw[10];

#Ball ~ Poisson[6.0];

TrueWeight(b) ~ UniformReal [0.0, 100.0];

BallDrawn(d) ~ UniformChoice({Ball b});

ObsWeight(d) {
  if BallDrawn(d) != null then
    ~UnivarGaussian[1](TrueWeight(BallDrawn(d)))
};

obs ObsWeight(Draw1) = 61.8;
obs ObsWeight(Draw2) = 64.4;
obs ObsWeight(Draw3) = 17.7;
obs ObsWeight(Draw4) = 81.8;
obs ObsWeight(Draw5) = 40.9;
obs ObsWeight(Draw6) = 81.9;
obs ObsWeight(Draw7) = 82.3;
obs ObsWeight(Draw8) = 82.9;
obs ObsWeight(Draw9) = 82.6;
obs ObsWeight(Draw10) = 60.8;

query TrueWeight(BallDrawn(Draw1));
query TrueWeight(BallDrawn(Draw2));
query TrueWeight(BallDrawn(Draw3));
query TrueWeight(BallDrawn(Draw4));
query TrueWeight(BallDrawn(Draw5));
query TrueWeight(BallDrawn(Draw6));
query TrueWeight(BallDrawn(Draw7));
query TrueWeight(BallDrawn(Draw8));
query TrueWeight(BallDrawn(Draw9));
query TrueWeight(BallDrawn(Draw10));
```

Figure 8: Example of selecting balls with replacement from an urn and measuring their weight

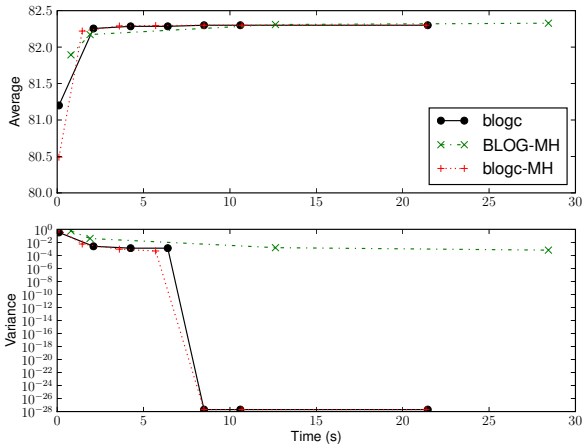

Figure 9: Balls with unknown weights

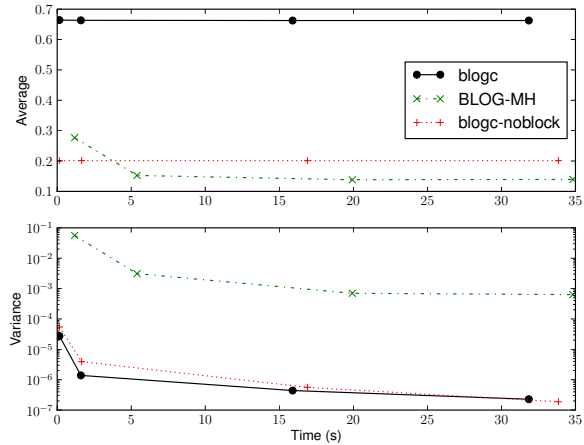

Figure 10: Results on the radar model

$WingType(Source(b1)) = Helicoper$. Given that $BladeFlash(b1) = true$ we expect this probability to be quite high. blogc converges to the true probability in less than a second. However, neither BLOG-MH nor blogc-noblock are able to come close to the true probability even after 30 seconds. This is explained by the fact that these two samplers are unable to directly sample the $WingType(a)$ variables. The fact that they are able to make any progress at all is due to the birth move which creates new aircraft for each WingType and samples their $RotorLen$ variable. Later, the move which resamples $Source(b)$ for each blip has the opportunity to select this new aircraft. These two moves thus compensate for the fact that the move which attempts to sample $WingType(a)$ is always rejected.

In follow-on work, we plan to demonstrate inference performance comparable to model-specific inference code for a number of widely used statistical models.

# 7 Conclusions

We have demonstrated a significant improvement in inference performance for models written in the BLOG language. Our Gibbs sampling algorithm for CBNs and our compiler techniques for generating efficient inference code are generally applicable to all open-universe stochastic languages.

### Acknowledgements

This work wouldn't have been possible without the considerable assistance provided by Brian Milch to make the models presented here work in BLOG-MH. Matthew Can provided a translation of the Alarm Bayes Net to BLOG. Finally, the first author wishes to thank his family for their boundless patience and support during this work.

### References

Andrieu, C., de Freitas, N., Doucet, A., & Jordan, M. I. (2003). An introduction to MCMC for machine learning. *Machine Learning*, *50*, 5–43.

Beinlich, I., Suermondt, G., Chavez, R., & Cooper, G. (1989). The alarm monitoring system: A case study with two probabilistic inference techniques for belief networks. *Proc. 2'nd European Conf. on AI and Medicine.* Springer-Verlag, Berlin.

Friedman, N., Goldszmidt, M., Heckerman, D., & Russell, S. (1997). Challenge: Where is the impact of Bayesian networks in learning? *IJCAI*.

Gelfand, A. E., & Smith, A. F. M. (1990). Sampling-based approaches to calculating marginal densities. *JASA*, *85*, 398–409.

Geman, S., & Geman, D. (1984). Stochastic relaxation, Gibbs distributions, and the Bayesian restoration of images. *IEEE Trans. on Pattern Analysis and Machine Intelligence*, *6*, 721–741.

Goodman, N., Mansinghka, V., Roy, D., Bonawitz, K., & Tenenbaum, J. (2008). Church: a language for generative models. *UAI*.

Milch, B., Marthi, B., Russell, S. J., Sontag, D., Ong, D. L., & Kolobov, A. (2005a). BLOG: Probabilistic models with unknown objects. *IJCAI* (pp. 1352–1359).

Milch, B., Marthi, B., Sontag, D., Russell, S., Ong, D. L., & Kolobov, A. (2005b). Approximate inference for infinite contingent Bayesian networks. *In Proc. 10th AISTATS* (pp. 238–245).

Milch, B., & Russell, S. (2006). General-purpose MCMC inference over relational structures. *Proceedings of the Proceedings of the Twenty-Second Conference Annual Conference on Uncertainty in Artificial Intelligence (UAI-06)* (pp. 349–358). Arlington, Virginia: AUAI Press.

Neal, R. M. (2000). Markov chain sampling methods for dirichlet process mixture models. *Journal of Computational and Graphical Statistics*, *9*, 249–265.

Spiegelhalter, D., Thomas, A., Best, N., & Gilks, W. (1996). *BUGS: Bayesian inference using gibbs sampling, version 0.50* (Technical Report).

Tait, P. (2009). *Introduction to radar target recognition*. The Institution of Engineering and Technology, United Kingdom.